\documentclass[revtex4]{article}
\usepackage[dvipdfm]{graphicx} 
\usepackage{amsmath,amssymb}
\usepackage{bm}
\usepackage{subfigure}
\usepackage{ascmac}
\usepackage{latexsym}
\usepackage{natbib}

\def\s|{\;\!|\;\!}   
\def\t|{\:\!|\:\!}  
\newtheorem{dfn}{Definition}[section]
\newtheorem{thm}{Theorem}[section]

\newtheorem{lem}{Lemma}[section]

\newtheorem{pos}{Postulate}[section]
\newtheorem{prf}{\textit{Proof.}}

\title{\bf{A Thermodynamical Approach for Probability Estimation}}
\author{Takashi Isozaki\\
  Sony Computer Science Laboratories, Inc.\\
  3-14-13 Higashigotanda Shinagawa-ku, 
  Tokyo 141-0022 Japan.\\
  isozaki@csl.sony.co.jp
}
\begin{document}
\maketitle
%
\begin{abstract}
The issue of discrete probability estimation for samples of small size is 
addressed in this study. The maximum likelihood method often suffers overfitting 
when insufficient data is available. 
Although the Bayesian approach can avoid overfitting by using prior distributions, 
it still has problems with objective analysis. In response to these drawbacks, 
a new theoretical framework based on thermodynamics, where energy and temperature 
are introduced, was developed. 
Entropy and likelihood are placed at the center of this method. 
The key principle of inference for probability mass functions is the minimum free energy, 
which is shown to unify the two principles of maximum likelihood and maximum entropy. 
Our method can robustly estimate probability functions from small size data. 
\end{abstract}

\section{Introduction}
A method for estimating probability of discrete random variables was developed. 
It is based on the key idea that statistical inference can be described by a combination of 
two frameworks, namely, thermodynamics and information theory. 
The roles of temperature and entropy in the method are paid special attention. 
In other words, \textit{heat} is introduced to statistical inference. 
This method, furthermore, has no free parameters, including temperature. 
The proposed method makes it possible to unify the maximum likelihood principle 
and the maximum entropy principle for statistical inference 
even from data of small sample sizes.

In recent times, the amount of various available data has been growing day by day. 
As a result, a large amount of data not only for one variable but for many variables can be obtained. 
Intuitively, getting conditional probabilities and joint probabilities can reduce the entropy 
of interested variables, which is guaranteed by information theoretic inequalities. 
Note that in this paper a capital letter such as $X$ denotes a random discrete variable, 
a non-capital letter such as $x$ denotes the special state of that variable, a bold capital letter 
denotes a set of variables, and a bold non-capital letter denoted configurations of that set. 
Here, we adopt Gibbs-Shannon entropy~\cite{Shannon} as entropy (we call this entropy 
Shannon entropy hereafter), defined as
\[
H(X)= -\sum_{x}P(x)\log P(x),
\]
where $P$ is a probability mass function. 
The inequalities are thus given as 
\[
H(X) \geq H(X \s| Y)
\]
and
\[
H(X)+H(Y) \geq H(X,Y),
\]
where the condition of the equality is independence between variables $X$ and $Y$~\cite{Shannon}. 
It is therefore preferable to use data generated from many variables 
because, obviously, highly predictive statistical inference requires low-entropy parameters. 
In discrete, many-variable systems, statistical estimation of conditional probabilities and 
joint probabilities needs exponentially large data 
because of combinatorial explosion of events in variables. 
Although the maximum likelihood (ML) principle and methods play significant roles in statistics 
and are regarded as the most-general principle in statistics, 
it is known that ML methods often suffer overfitting and 
that they are ineffective in cases such as insufficiently large data size 
in relation to number of parameters.

The situation that ML methods often suffer overfitting in multivariate statistical analysis 
with many parameters seems to make Bayesian statistics more and more attractive 
from the viewpoint of avoiding overfitting. 
Bayesian statistics incorporates background knowledge, which compensates for shortage of data size 
and increasingly becomes popular in natural science including physics~\cite{BayesInPhysDose}. 
It can be said that Bayesian statistics adds prior imaginary frequency of events to real one. 
Furthermore, even in the cases of no available prior knowledge or the case of public analysis which needs 
to preclude prior knowledge for avoiding generation of unnecessary bias, Bayesian statistics can 
reduce overfitting by means of \textit{noninformative priors} (e.g.,~\citet{KassWassermanSelection}). 
For example, in discrete random-variable systems, Bayesian statistics usually uses Dirichlet distributions, 
which always have parameters. 
The parameters of Dirichlet prior distributions are often interpreted as prior samples. 
When those parameters are uniform and express no special prior knowledge, they can 
\textit{increase entropy of ML estimators} and thereby make estimated probabilities 
more robust than those obtained from ML estimation. 
This feature of those parameters can be regarded as one to generalize 
\textit{the principle of insufficient reason} proposed by Laplace.

Although noninformative priors have been widely used, 
they still have some 
problems ~\cite{KassWassermanSelection, NonInformativeDialogue, BayesianChoice}. 
For example, Jeffreys' priors~\cite{JeffreysBook}, 
which are the most widely accepted noninformative priors, 
do not satisfy axioms of probability and are thus said to be improper distributions. 
In addition, even the posterior can be still improper~\cite{KassWassermanSelection}. 
As for statistical inference, it is thus probably reasonable to explore another principle 
or theoretical framework. 

A theoretical framework for estimating probability of discrete variables 
used in objective analysis is proposed in the following. 
The new framework keeps the good characteristics of Bayesian statistics: 
increasing entropy obtained from ML estimators in the case with no prior knowledge. 
Entropy is regarded as representing the uncertainty of information. 
Accordingly, the entropy in the case of insufficient data size should be larger than 
that obtained from hidden true distributions, because the smaller the data size is, 
the higher uncertainty becomes. 
That is, we consider that \textit{entropy consists of uncertainty due to limited sample size 
and uncertainty due to true probability distributions}. 
It is therefore proposed that probability estimation should be regarded as searching 
for the optimal value of entropy according to available data size and data property. 
However, regarding limited sample size (which increases uncertainty), 
neither a satisfactory principle nor a method for optimally estimating entropy exists.
\citet{Jaynes57I} proposed a method based on the maximum entropy principle, 
which has been used in some domains 
including physics (e.g., \cite{MEtoNLP,ME_Huscroft2000,MEpriorPRE04}). 
Our method utilizes information of frequencies like ML methods in addition to 
ME principle, both of which cooperate to modify over biases due to small samples, 
while Jaynes' method does not. 
We regard the difference is an essential one between the both methods. 

The theoretical framework on which the proposed probability estimation method 
is based consists of and unifies two well-known principles. 
The first is the maximum likelihood (ML) principle which states that 
the best estimators should most duplicate data and which is 
very effective in the case of sufficiently large data size. 
The second is the maximum entropy (ME) principle, which states that 
no bias should be applied to particular internal states of variables, 
within some constraints, as far as possible. 
However, each principle is contrary to the other because the expectation values 
of minus-log likelihoods are the same as empirical entropy. 
In clear contrast to the ME principle, it is intuitively obvious that obtaining 
the estimator with the lowest entropy and the highest likelihood is preferable 
in the case of sufficiently large data.

Given the above-described conflict between ML and ME principles, 
it is necessary to devise a method 
for analyzing samples with insufficient data size. 
It seems natural that there is a balancing value 
between of the entropies given by the ML and ME principles. 
If it is assumed that such a balancing value exists, it is necessary 
to find the optimal point between contrary principles in statistical inference. 

In a branch of natural science, namely, thermodynamics, 
there is an analogy with the above-described trade-off between 
the ML and ME principles~\cite{Kittel, CallenThermodynamics}. 
In thermodynamics, nature selects the state that achieves a balance between 
the minimum energy state and the maximum entropy state at a finite temperature. 
It is assumed here that this analogy applies to statistical inference in discrete variables. 
Consequently, \textit{temperature}, which plays the role of a unit of measure, is introduced. 
This approach is an extended one from our preceding works~\cite{isozakiICTAI, Isozaki_IJAIT}, 
in which temperature was represented by an artificial model containing 
a free hyperparameter. 
In the present work, temperature is entirely redefined in a new method with \textit{no} 
free hyperparameters. 
According to our proposed method, 
a new interpretation of probability estimated from data 
is presented, which is neither frequentism~\cite{HajekArguments1997} nor Bayesianism.

In the machine learning domain, some similar methods to ours have been 
proposed~\cite{Pereira93ACL,UedaDAEM,Hofmann99UAI, YannLeCun_AISTATS05,KWataVFEforBN}. 
Nevertheless, many studies that have applied free energy to statistical science 
have not included temperature or treated as 
a controlled parameter, fixed parameter or a free parameter, 
apparently because of the lack of clarity of its meaning in data science. 
In regard to the existing researches, therefore, we consider that 
the potentials of free energies are not well extracted. 
Similar methods in context of robust estimation, 
in which a free parameter is introduced in a similar fashion, 
have also been investigated~\cite{Windham1995Robust, BasuBetaLikelihood,JonesBiometrika2001}, 
where how to determine the free parameter for small samples still remains all the same.


This paper is organized as follows. In the next section, 
the basic theory based on thermodynamics is explained. 
The proposed ``probability estimation method'' is introduced in Section~\ref{PE}, 
where estimation methods for joint probabilities and conditional probabilities are also proposed. 
Section~\ref{Example} presents 
the results of experiments using the probability estimation method. 
The relationships between our method and classical/Bayesian statistics 
are discussed in Section~\ref{Discussion}. 
Section~\ref{Conclusion} concludes this study. 

\section{Basic theory}
In constructing an estimation method of finite discrete probability 
distributions from samples with finite size, 
we utilize both Shannon entropy and likelihood. 
However, a new principle is needed for combining the two concepts; accordingly, 
we assume the principle to do it is in the thermodynamical framework. 
In the following, therefore, entropy, energy, temperature, and Helmholtz 
free energy are defined for the purpose. 
Necessary postulates for constructing the method and its properties 
are then described. 
Hereafter, multivariate random systems are treated without 
any prior knowledge. 
All probability distributions are assumed to be discrete variables, 
and samples are assumed to be i.i.d. data. 
An extension to the case with available prior knowledge 
is discussed in section~\ref{EX}. 

\subsection{Definitions and postulates}
\subsubsection{Entropy}
\begin{dfn}
[Entropy]
\label{DefEntropy}
The entropy $H(X)$ of a discrete random variable $X$ is defined, according to~\citet{Shannon}, as 
\begin{equation}
H(X):= -\sum_{x}P(x)\log P(x).\label{Entropy}
\end{equation}
If $P(X)$ is an estimated probability function, $H(X)$ is also an estimated function of $P(X)$. 
Entropy is also denoted as $H(P(X))$ or $H(P)$ in order to make it clear which distributions are used. 
\end{dfn}
Joint entropy $H(\bm{X})$ of multivariate systems and conditional entropy $H(X \s| \bm{Y})$ 
are defined as follows~\cite{Shannon,CoverThomas}: 
\begin{equation}
H(\bm{X}):= -\sum_{\bm{x}}P(\bm{x}) \log P(\bm{x})
\end{equation}
and
\begin{equation}
H(X \s| \bm{Y}):= -\sum_{x,\bm{y}}P(x,\bm{y})
\log P(x \s|\bm{y}).
\end{equation}
It should be noted that probability mass function $P(x)$ and entropy $H$ are 
quantities that should be estimated from data in this study. 
Accordingly, it should be emphasized that the entropy has two aspects of uncertainty: 
The first is the uncertainty that each true probability distribution peculiarly 
has; the second is that which comes from finiteness of available data. 
To the author's knowledge, the second aspect of entropy has not been specifically discussed. 
Accordingly, we introduce a mechanism to estimate \textit{the optimal uncertainty under 
given finite available data}.

\subsubsection{Energy}
The (internal) energy of a probability system is defined as follows. 
First, a distance-like quantity between two distributions is defined 
in the usual way as follows. 
\begin{dfn}
[Kullback-Leibler divergence]
The Kullback-Leibler (KL) divergence \cite{KL} between two distributions of a 
random variable $X$, i.e., $P(X)$ and $Q(X)$, is given as follows:
\begin{equation}
D(P(X)\,||\,Q(X))
:=\sum_{x}P(x)\log \frac
{P(x)}{Q(x)}.
\end{equation}
\end{dfn}
For multivariate systems, $D(P(\bm{X})\,||\,Q(\bm{X}))$ can be 
defined in the same manner. 
Conditional KL divergence is defined as 
\begin{equation}
D(P(X\s|\bm{Y})\,||\,Q(X\s|\bm{Y})):=
\sum_{x,\bm{y}}P(x,\bm{y})\frac{P(x \s| \bm{y})}{Q(x \s| \bm{y})}.
\end{equation}

Next, the cross entropy is defined, which is also useful 
to represent the energy. 
\begin{dfn}
[Cross entropy]
\label{DefCrossEnt}
The cross entropy of discrete random variable $X$ between 
probability distributions $P(X)$ and $Q(X)$, i.e., $H(P(X),Q(X))$, 
is defined as 
\begin{equation}
H(P(X),Q(X)):= -\sum_{x}P(x)\log Q(x).\label{CrossEnt}
\end{equation}
\end{dfn}
Cross entropy is also denoted as $H(P,Q)$. 
The following relationship between KL divergence and cross entropy 
is easily derived: 
\begin{equation}
D(P(X)\,||\,Q(X))+H(P(X))=H(P(X),Q(X)).\label{KL-CE}
\end{equation}
According to Jensen's inequality, $D(P\,||\,Q) \geq 0$~\cite{CoverThomas} 
and $H(P,Q) \geq H(P)$.

Empirical distribution functions are defined in a usual way. 
\begin{dfn}
[Empirical distributions]
It is assumed that there are $N$ samples of random variable 
$X$: $\{y^{(1)},...,y^{(N)} \}$. 
An empirical distribution of $X$, $\tilde{P}(X)$, is defined as 
\begin{equation}
\tilde{P}(X=x)=\frac{1}{N}\sum_{i=1}^{N}\delta(x-y^{(i)}),
\end{equation}
where $\delta(x-y)=1$ if $x=y$ and $\delta(x-y)=0$ if $x \neq y$.
\end{dfn}
$\tilde{P}(X)$ is relative frequency, that is, a maximum likelihood (ML) 
estimator, which is denoted by $\hat{P}(X)$.

\begin{dfn}
[Information energy]
\label{DefEnergy}
(Internal) energy is defined by using Kullback-Leibler divergence as a 
distortion between the distribution of a target and an empirical 
distribution: 
\begin{equation}
U_{0}(X):=D(P_{1}(X)\,||\,P_{2}(X)),\label{EnergyKL}
\end{equation}
where $P_{1}(X)$ denotes a target mass function to be estimated, and $P_{2}(X)$ 
denotes an empirical function or the ML estimator. 
Cross entropy can also be used as an alternative of the KL divergence: 
\begin{equation}
U(X):=H(P_{1}(X),\,P_{2}(X)).\label{EnergyCE}
\end{equation}
$U_{0}$ and $U$ are, hereafter, both called ``information energy''. 
\end{dfn}
It is noteworthy that minimizing $U_{0}$ for probability estimation 
corresponds to the ML principle. 

Self-information energy of functions of $X$ or $\bm{X}$, namely, $\epsilon$, 
are defined as 
\begin{equation}
\epsilon(X) := - \log \tilde{P}(X),\label{SelfEneDef}
\end{equation}
where $\tilde{P}(X)$ is the empirical distribution, and $\epsilon(X)$ can be 
defined for any states $x$ of $X$ when $\tilde{P}(x) > 0$. 
Practically, the ML estimator, $\hat{P}(X)$, can be used as an alternative to 
$\tilde{P}(X)$. That is, the following equation is used: 
\begin{equation}
\epsilon(X) := - \log \hat{P}(X). \label{SelfEne}
\end{equation}
This equation indicates that self-information energy denotes 
minus maximum-log likelihood.

\subsubsection{Temperature}
Inverse temperature, $\beta_{0}$, is introduced as one of the most significant 
quantities for statistical inference with finite data size. 
$\beta_{0}$ is used instead of physical temperature, often denoted as $T$, 
and is simply called ``temperature'' hereafter. 
Temperature is regarded as a bridge between thermodynamics and statistical inference. 
As described in our preliminary work~\cite{isozakiICTAI,Isozaki_IJAIT}, 
by introducing a thermodynamical framework for statistical inference, 
fluctuation due to finiteness of available data size 
can be regarded as thermal fluctuation. 
In the following, this philosophy is applied to define temperature 
in this paper for constructing a probability estimation method.

Fluctuation which data have can be denoted by the distortion between the ML 
estimator in currently available data size $n$ and the probability function 
estimated from the new framework by using $n-1$ 
data (which do not include the $n$th data). 
The fluctuation is related to the temperature as a unit of measure. 
$P_{i}(X)$ is first defined as a new estimator obtained from $i$ data, 
and averaged estimator $P_{n}^{G}(X)$, which denotes the geometric mean 
for data size $n\ (\ge 0)$, is defined as follows: 
\begin{equation}
P_{n}^{G}(X):= \left( \prod_{i=0}^{n}
P_{i}(X) \right) ^{\frac{1}{n+1}},
\label{PGA}
\end{equation}
where $P_{0}^{G}(X):=P_{0}(X):=1/|X|$ is defined as a uniform function, 
in which $|X|$ denotes the number of elements in the range of $X$. 
This definition for $n=0$ corresponds to the ME principle. 
It follows that the distortion is denoted as KL divergence, and 
the divergence is connected to the temperature. 
\begin{dfn}
[Data temperature]
\label{DefTemp}
For a natural-number data size, i.e., $n \ge 1$, 
the inverse temperature of a random variable $X$, namely, 
$\beta_{0}(X)$, is defined as 
\begin{eqnarray}
\beta_{0}(X)&:=&
1/D(P_{n-1}^{G}(X)\,||\,\hat{P}(X))\nonumber\\
&=&1/\sum_{x_{k}}P_{n-1}^{G}(x_{k})
\log \frac{P_{n-1}^{G}(x_{k})}{\hat{P}(x_{k})}
\label{TempFluctuation}
\end{eqnarray}
for $D(P_{n-1}^{G}(X)\,||\,\hat{P}(X)) \ne 0$, 
where $P_{n-1}^{G}(X)$ is defined by Equation (\ref{PGA}). 
$\beta_{0}(X):=0$ for data size $m = 0$, and $\beta_{0}(\bm{X})$ 
of multivariate systems, i.e., $\bm{X}$, can be defined in the same way. 
Note that variables $X$ and $\bm{X}$ are often omitted 
if they can be clearly recognized. 
\end{dfn}
We call $\beta_{0}$ ``data temperature''. 
According to this definition, 
it can be assumed that $0 < \beta_{0} < \infty$ for $n>0$. 
It will be seen that this positivity of temperature has consistency with 
a postulate described in~\ref{postulates} and Equation (\ref{DefUsualBeta}). 
In general, $\beta_{0}$ becomes large, that is, the system approaches 
low-temperature state, as available data size grows 
because the fluctuation of data becomes small. 

Normalized data temperature in statistical inference, which 
improves tractability of data temperature in mathematical formulas, 
is defined as follows: 
\begin{dfn}
[Data temperature II] 
For a natural-number data size, i.e., $n \ge 1$, temperature of 
a random variable, $\beta(X)$, is defined as 
\begin{equation}
\beta(X):=\frac{\beta_{0}(X)}{1+\beta_{0}(X)}.\label{defB}
\end{equation}
$\beta(\bm{X})$ is defined in the same way, 
and the variable(s) name is often omitted. According to this definition, 
$0 < \beta < 1$ for $n>0$. $\beta:=0$ for data size $m=0$. 
Hereafter, both $\beta$ and $\beta_{0}$ compatibly are used, and both are called 
``data temperature'' when there is in no danger of confusion. 
\end{dfn}

\subsubsection{Helmholtz free energy}
Helmholtz free energy, which plays a significant role in 
the method we present, is introduced next. 
\begin{dfn}
[Helmholtz free energy] 
Helmholtz free energy, $F$ for $X$, is defined by using information 
energy, $U_{0}(X)$, Shannon entropy, $H(X)$, and data temperature 
$\beta_{0}(X)$, as follows: 
\begin{equation}
F(X):=U_{0}(X)-\frac{H(X)}{\beta_{0}(X)}.\label{defF}
\end{equation}
Free energy can be equivalently rewritten using $\beta$ and cross entropy $U$, 
instead of $\beta_{0}$ and $U_{0}$, as
\begin{equation}
F(X):=U(X)-\frac{H(X)}{\beta(X)}.\label{defF2}
\end{equation}
For multivariate systems, $F(\bm{X})$ can be defined in the same way. 
\end{dfn}
The second terms of right-hand sides  
in Equations (\ref{defF}) and (\ref{defF2}) represent thermal energy 
in thermodynamics. 
It follows that the concept of \textit{heat} is explicitly introduced 
in statistical inference.

\subsubsection{Postulates}\label{postulates}
To assure the positivity of temperature, the following postulate, 
which will be needed in~\ref{UDT}, is first assumed. 
\begin{pos}\label{EntInc}
Entropy is differentiable and is a monotonically increasing function of energy. 
The coefficient of the partial derivative for the energy 
takes positive values as follows:
\begin{equation}
\frac{\partial H}{\partial U_{0}} > 0, 
\label{BetaHU}
\end{equation}
where $U_{0}$ denotes the energy represented by the KL divergence and
\begin{equation}
\frac{\partial H}{\partial U} > 0, 
\label{BetaHU2}
\end{equation}
where $U$ denotes the energy by the cross entropy. 
\end{pos}
The key principle of our method is assumed as follows. At the large sample limit, 
$U_{0}\rightarrow 0$ is reasonable in accordance with the ML principle, 
while it is reasonable that $H$ takes maximum values at the small sample limit 
in accordance with the maximum entropy (ME) principle. 
For a finite sample size, it is thus reasonable that estimators of probabilities 
take the values that balance both principles in accordance with the data size 
and the true hidden intrinsic entropies. 
We postulate that the minimum (Helmholtz) free energy principle determines 
the optimal balance. 
\begin{pos}
[Minimum free energy principle]
\label{P-MaxEnt}
Probability mass functions that are estimated from data 
are such as to minimize the Helmholtz free energy. 
\end{pos}
We call the principle MFE principle. 

The two above-stated postulates are all that is needed for the framework of our proposed method. 
It is noteworthy that these postulates are parts of thermodynamics~\cite{CallenThermodynamics}, 
implying that our method of statistical inference is fully based on 
the framework of thermodynamics theory (except for the interpretation of the entropy, 
for which Shannon entropy is adopted). 

Our new method developed here selects a probability mass function that maximizes entropy as far as possible 
according to the minimum free energy principle, while the maximum entropy method of~\citet{Jaynes57I} 
maximizes the entropy subject to his adopted another constraint. 
The new method even corrects the bias generated from the limited size samples, which is a major difference 
compared to Jaynes' method. We consider that the difference arises from the theoretical ground, 
which of our method is thermodynamics with Shannon entropy while which of Jaynes is only information theory. 

The basis of our method for inference is described by using Shannon entropy and introducing 
``information energy'', ``data temperature'', and the minimum free energy (MFE) principle. 
When we estimate probability functions from finite data, its purpose is getting effective 
\textit{information} from data for recognizing truth and/or predicting future events. 
From this viewpoint, with a finite data size, it is reasonable to select a probability function 
that explains data to some extent but has some additive uncertainty due to having limited samples. 
MFE principle, thus, unifies ML and ME principles, and thermodynamics also has 
a similar relationship: MFE principle unifies minimum (internal) energy principle 
and maximum entropy principle~\cite{CallenThermodynamics}. 

\section{Probability estimation}
\label{PE}
A probability-estimation method based on the theory described in the previous section 
is formalized in the following. The estimation is based on the MFE principle. 
Multinomial distributions are used for discrete random variables in the usual manner. 

\subsection{Probability estimation method}
\label{PESV}
The entropy of a variable, $X$, is first defined by Equation (\ref{Entropy}). 
Information energy is then defined by Equation (\ref{EnergyKL}), probability $P(X)$, 
and the empirical distribution ($\tilde{P}(X)$) as 
\begin{equation}
U_{0}(X):=D(P(X)\,||\,\tilde{P}(X))
=\sum_{x} P(X)\log \frac{P(X)}{\tilde{P}(X)}.
\label{EnergyEstimation}
\end{equation}
$\tilde{P}(X)$ is replaced by the ML estimator $\hat{P}(X)$ because $\hat{P}(X)$ denotes 
relative frequency, which is the same as $\tilde{P}(X)$ 
for binomial or multinomial distributions under the condition of \textit{i.i.d}. 

According to the MFE principle, 
probability estimator $P(X)$ with $\beta_{0}$ or $\beta$ is estimated by minimizing $F$ 
under the constraint $\sum_{x}P(x)=1$. 
It is therefore solved by using Lagrange multipliers. 
Free energy $F$ is written in the following form with Shannon entropy and information energy:
\begin{equation}
F=U_{0}-\frac{1}{\ \beta_{0}\ }H.
\end{equation}
$\beta$ can be used as the alternative to $\beta_{0}$; accordingly, free energy $F$ 
can be rewritten by using cross entropy $U$ as
\begin{equation}
F=U-\frac{1}{\ \beta \ }H. \label{FUHBrelation}
\end{equation}
It follows that $\{U,\beta\}$ can be used instead of $\{U_{0},\beta_{0}\}$. 
The Lagrangian $L$ is expressed as 
\begin{eqnarray}
L&=&F+\lambda \left( \sum_{x} P(x)-1 \right)\nonumber\\
&=&\frac{1}{\ \beta\ }\sum_{x}
P(x) \log P(x) -\sum_{x}P(x) \log \hat{P}(x)\nonumber\\
&\ &+ \lambda \left( \sum_{x} P(x)-1 \right),
\end{eqnarray}
where $\lambda$ is the Lagrange multiplier. 
In relation to that expression, if $\beta_{0} \rightarrow 0$, 
then $\beta \rightarrow 0$ (high-temperature limit); if $\beta_{0} \rightarrow \infty$, 
then $\beta \rightarrow 1$ (low-temperature limit). 
The solution $P(X)$ is thus derived from the following equation: 
$\partial L / \partial P(x)=0$. 
The estimated probability, $P(X)$, is therefore expressed in the form of 
the canonical distributions, which is also called Gibbs distributions. 
The distribution as the solution is well known in statistical physics as 
\begin{eqnarray}
P(x)&=&\frac{\exp(-\beta(-\log \hat{P}(x)))}
{\sum_{x'}\exp(-\beta(-\log \hat{P}(x')))}\label{Boltzmann}\\
&=&\frac{\exp(-\beta \epsilon(x))}
{\sum_{x'}\exp(-\beta \epsilon(x'))},
\end{eqnarray}
where $\epsilon(x)$ as expressed in Equation (\ref{SelfEne}) is used. 
Practically, the following equivalent form is used:
\begin{equation}
P(x)=\frac{[\hat{P}(x)]^{\beta}}{\sum_{x'}[\hat{P}(x')]^{\beta}},
\label{P-Beta-prac}
\end{equation}
where $\beta$ can be determined, without any free parameters, by using 
Equations (\ref{PGA}), (\ref{TempFluctuation}), (\ref{defB}), and (\ref{P-Beta-prac}). 
For data size $n=0$, the estimator is defined such that $P(x)=1/|X|$, 
where $|X|$ denotes the number of elements in the range of $X$. 
Note that the proposed method 
has consistency with the ME principle, 
at high temperature limit, 
where $\text{min}\,F \approx \text{max}\,(H/\beta)$, and 
with ML principle, at low temperature limit, 
where $\text{min}\,F \approx \text{min}\,U$.

For conditional probability, conditional entropy $H(X \,|\, \bm{Y})$ and 
conditional KL divergence $D(P(X \,| \bm{Y})\,||\, \hat{P}(X \,| \bm{Y}))$ 
or conditional cross entropy are used. 
$\beta$ is defined as 
$\beta(X \,| \bm{Y}):=\beta_{0}(X \,| \bm{Y})/ (\beta_{0}(X \,| \bm{Y})+1)$. 
The formula for estimating conditional probabilities is 
therefore obtained in the following form:
\begin{equation}
P(x \s| \bm{y})=\frac{\exp(-\beta (-\log \hat{P}(x\s|\bm{y})))}
{\sum_{x'}\exp(-\beta (-\log \hat{P}(x'\s|\bm{y})))}.
\label{CondP}
\end{equation}
For conditional data size $n=0$ given $\bm{Y}=\bm{y}$, $P(x \s| \bm{y})=1/|X|$ 
for any $\bm{y}$ in the same manner as $P(x)$ given in section~\ref{PESV}. 

Joint probability can be calculated by using Equations (\ref{Boltzmann}) 
and (\ref{CondP}) and the definite relation: $P(X,Y)=P(X|Y)P(Y)$. 
In general, it is calculated using decomposition rules such that 
\begin{eqnarray*}
&&P(X_{1},X_{2},\dots,X_{n})=\nonumber\\
&&\ \ \ P(X_{n}\,|\,X_{n-1},\dots,X_{2},X_{1}) 
\dots P(X_{2}\,|\,X_{1})P(X_{1}). 
\end{eqnarray*}

Partition functions similar to statistical mechanics are introduced for convenience. 
By using ``data temperature'' $\beta$, free energy $F$ is expressed in the same 
form as that in statistical mechanics: 
\begin{eqnarray}
F&=&U-\frac{1}{\ \beta \ }H\nonumber\\
&=&-\frac{1}{\,\beta\,}\log Z,\label{F-Z}
\end{eqnarray}
where $Z$ is the partition function, which is well known in statistical mechanics, 
defined for single or multivariate probabilities as  
\begin{equation}
Z(\bm{X})=\sum_{\bm{x}}[\hat{P}(\bm{x})]^{\beta},
\end{equation}
and for conditional probabilities as 
\begin{equation}
Z(X\s|\bm{Y})=\sum_{x}[\hat{P}(x\s|\bm{Y})]^{\beta}.
\end{equation}
Consequently, when the thermodynamical framework and Shannon entropy are assumed, 
the partition-function formula of statistics common to statistical mechanics can be derived.

A significant feature of data temperature is proved as follows. 
The lemma needed for this proof is stated as follows. 
\begin{lem}\label{PnGconv}
$P_{i}(X)$ is denoted as the canonical distribution estimated by Equation (\ref{Boltzmann}) 
from $i$ data. 
For data size $n \rightarrow \infty$, $P_{n}^{G}(x)$, defined by Equation (\ref{PGA}), 
converges to a definite value $P^{G}(x)$ when $0 < P_{i}(x)$ for integers $i$ 
such that $i \geq 0$ and any state $x$. 
\end{lem}
\begin{prf}
\begin{eqnarray}
& &\log P_{n}^{G}(x)-\log P_{n-1}^{G}(x)\nonumber\\
&=&\frac{1}{n+1}\sum_{i=0}^{n}
\log P_{i}(x)-\frac{1}{n}\sum_{i=0}^{n-1}\log P_{i}(x)\nonumber\\
&=&\left(\frac{n}{n+1}-1\right)\log P_{n-1}^{G}
+\frac{1}{n+1}\log P_{n}(x).
\label{PnGconv1}
\end{eqnarray}
Because $0 < P_{n-1}^{G}(x)$ and $0 < P_{n}(x)$ and then $\log P_{n-1}^{G}(x)$ 
and $\log P_{n}(x)$ are definite values, 
if $n \rightarrow \infty$, both terms on the right-hand side of Equation (\ref{PnGconv1}) 
converge to $0$. 
Therefore, $P_{n}^{G}(x) \rightarrow P^{G}(x)$ because log functions are single-valued functions. 
\end{prf}
\begin{thm}\label{ThDT}
At the asymptotic limit (i.e. large sample limit), data temperature $\beta$ converges to $1$ 
when $0 < P_{i}(x)$ for integers $i$ such that $i \geq 0$ and any state $x$, 
where $P_{i}(x)$ is denoted as the canonical distribution estimated 
by Equation (\ref{Boltzmann}) from $i$ data. 
\end{thm}
\begin{prf}
According to Lemma~\ref{PnGconv}, $P_{n}^{G}(x) \rightarrow P^{G}(x)$ 
at the limit $n \rightarrow \infty$, where $P^{G}(x)$ is a definite value for any $x$. 
$P^{G}(x)$ is also a definite value: $P_{n}(x)$, where $P_{n}(x)$ is 
an estimated value given by Equation (\ref{Boltzmann}) using $n$ data, 
because $\log P^{G}(x)$ is a mean value of $\log P_{i}(x)$. 
$\beta$ thus converges to a definite value. 
Meanwhile, ML estimator $\hat{P}(x)$ converges to true distribution $P_{t}(x)$ 
due to the consistency of ML estimators. $P_{n}(x)$ at $n \rightarrow \infty$ is 
denoted as $P(x)$. Therefore, at $n \rightarrow \infty$, 
\begin{eqnarray}
\frac{1}{\beta_{0}}=\frac{1-\beta}{\beta}
&=&D(P(X)\,||\,P_{t}(X))\nonumber\\
&=&\sum_{x} \frac{[P_{t}(x)]^{\beta}}{Z} \log 
\frac{[P_{t}(x)]^{\beta-1}}{Z} \label{infnident}
\end{eqnarray}
for $0< P(x)$ and any state $x$. 
This identity (\ref{infnident}) needs $\beta \rightarrow 0$ or $\beta \rightarrow 1$ 
in order that $[P_{t}(x)]^{\beta}$ or $[P_{t}(x)]^{\beta -1}$ is a constant 
for any probability distributions $P_{t}(x)$. 
However, $\beta \rightarrow 0$ does not satisfy Equation (\ref{infnident}), 
while $\beta \rightarrow 1$ does satisfy the equation. 
Accordingly, $\beta \rightarrow 1$ is the asymptotic limit. 
\end{prf}

According to Theorem~\ref{ThDT}, the more data is obtained, the more $\beta$ 
approaches $1$ and the more the estimator approaches the ML estimator. 
Therefore, it is noticeable that the new estimator has the same preferable 
asymptotic properties as the ML has, which are consistency and 
asymptotic efficiency. 

For insufficient data size, $\beta_{0}$ is small by definition, so $\beta$ is also small. 
Adequate estimators, which are automatically adjusted to available data, 
can therefore be obtained. In other words, free energy is dominated 
by the second term of Equation (\ref{FUHBrelation}) 
when sufficient data is not available, because uncertainty is large due to 
shortage of evidence. 
In contrast, it is dominated by the first term when sufficient data is available, 
because uncertainty is small due to a large size of data. 
We call the proposed method \textit{MFEE}, which we abbreviate ``MFE estimation'' as. 


\subsection{Interpretation of probability and estimated information}
MFEE provides a new interpretation of probability instead of 
frequentism or Bayesianism. 
Frequentism is based on counting occurrence of events (e.g.~\cite{HajekArguments1997}) 
and Bayesianism is based on subjectivity or combination of counting prior imaginary 
and real occurrences of events. 
It can be regarded that the Bayesian approach extends frequentism to that including (prior) 
imaginary  counting of events. 
The thermodynamical estimation method stated in this section is based on the concept of 
optimal uncertainty, which consists of counting events and temperature. 
In MFEE, probability can be regarded as the degree of uncertainty according to 
the MFE principle optimizing uncertainty in a reflection of quality and quantity of the data. 
It can therefore be regarded as a new interpretation of probability in real-world applications.

When the optimal entropy is obtained by using the MFE principle, the optimal negative entropy 
represents the optimally estimated effective information, which is defined as $EI$ 
(denoting ``effective information''), 
for given data as follows: 
\begin{eqnarray*}
EI(X)
:=\sum_{x}\frac{[\hat{P}(x)]^{\beta}}{Z}
\log \frac{[\hat{P}(x)]^{\beta}}{Z}. 
\end{eqnarray*}
The averaged log likelihood is therefore a large sample approximation of $EI$.

\subsection{Some characteristic properties of MFEE}
The canonical distribution derived from the MFE principle can provide 
some characteristic properties of MFEE. 
The following notations are defined. 
Probability mass functions, such as $P_{k}$, have discrete states that are 
denoted as index $k$. 
The ML estimator is denoted by $\hat{P}_{k}$. $\{\beta, U\}$ is used instead 
of $\{\beta_{0},U_{0}\}$. 

\subsubsection{Relation to usual definition of temperature}\label{UDT}
In thermodynamics, temperature $\beta$ is usually defined 
as~\cite{CallenThermodynamics,Kittel}
\begin{equation}
\beta := \frac{\partial H}{\partial U}.
\label{DefUsualBeta}
\end{equation}
However, if the canonical distribution, which has the form of Equation (\ref{Boltzmann}), 
is used, the usual definition of $\beta$ is automatically satisfied as follows. 

\begin{lem}
$H=\beta U + logZ$ under the MFE condition, 
where $H$, $U$, $\beta$, and $Z$ denote entropy, information energy, data temperature, 
and partition function. 
\end{lem}
\begin{prf}
Since probability mass function $P_{k}$ has a canonical form under the MFE condition, 
it follows that
\begin{eqnarray}
H&=& -\sum_{k}P_{k} \log P_{k}=-\sum_{k}\frac{\hat{P}_{k}^{\beta}}{Z}
\log \frac{\hat{P}_{k}^{\beta}}{Z}\nonumber\\
&=& \beta \cdot 
\left( -\sum_{k}\frac{\hat{P}_{k}^{\beta}}{Z} \log \hat{P}_{k} \right)
+\left( \sum_{k}\frac{\hat{P}_{k}^{\beta}}{Z}\right) \cdot \log Z
\nonumber\\
&=& \beta U + \log Z,\label{HUBZ}
\end{eqnarray}
where cross entropy is used as $U$. 
\end{prf}

\begin{thm}
Equation (\ref{DefUsualBeta}) is automatically satisfied under the MFE principle. 
\end{thm}
\begin{prf}
Differentiating partially with respect to $U$ of both sides of Equation (\ref{HUBZ}) gives 
\begin{equation}
\frac{\partial H}{\partial U} = \beta 
+ U \frac{\partial \beta}{\partial U} 
+ \frac{\partial}{\partial U} \log Z.
\end{equation}
\begin{equation}
\frac{\partial}{\partial U} \log Z = 
\frac{\partial \beta}{\partial U}
\frac{\partial}{\partial \beta} \log Z.
\end{equation}
\begin{eqnarray}
\frac{\partial}{\partial \beta} \log Z
&=& \frac{1}{\ Z \ }\frac{\partial}{\partial \beta} 
(\sum_{k} \hat{P}_{k}^{\beta})
=\frac{1}{\ Z \ } \sum_{k} \hat{P}_{k}^{\beta}
\log \hat{P}_{k}\nonumber\\
&=& \sum_{k} \frac{\hat{P}_{k}^{\beta}}{Z} \log \hat{P}_{k}
= - U. \nonumber
\end{eqnarray}
It follows that
\[
\frac{\partial H}{\partial U}=\beta.
\]
The theorem is therefore proved. 
\end{prf}
In the same way, it can be proved that $\beta_{0}=\partial H / \partial U_{0}$.

\subsubsection{Energy fluctuation}
In statistical mechanics, energy fluctuations 
$\langle \epsilon ^{2}\rangle - \langle \epsilon \rangle^{2}$ are shown to 
have the following relation, where $\langle \rangle$ denotes 
an expectation value in respect to the canonical distributions. 
\begin{equation}
\langle \epsilon ^{2}\rangle - \langle \epsilon \rangle^{2}
= -\frac{\partial U}{\partial \beta}.
\label{EnergyFluctuation}
\end{equation}
In regard to energy defined in MFEE, namely, 
Equation (\ref{SelfEne}), the same relation as that shown here is satisfied. 
We use the following equation: 
\begin{eqnarray}
U=-\sum_{k}\frac{\hat{P}_{k}^{\beta}}{Z}\log \hat{P}_{k},
\end{eqnarray}
where $U$ denotes cross entropy. 
\begin{eqnarray}
-\frac{\partial U}{\partial \beta}
&=& \sum_{k}(\log \hat{P}_{k})
\left\{\frac{1}{Z}\hat{P}_{k}^{\beta} \log \hat{P}_{k}
-\left(\frac{1}{Z^{2}}\right)\sum_{m}
\frac{\partial \hat{P}_{m}^{\beta}}{\partial \beta}\right\}\nonumber\\
&=& \sum_{k}(\log \hat{P}_{k})
\left \{ \frac{\hat{P}_{k}^{\beta}}{Z}\log \hat{P}_{k}
-\frac{\hat{P}_{k}^{\beta}}{Z}\sum_{m}
\frac{\hat{P}_{m}^{\beta}}{Z}\log \hat{P}_{m}\right \}\nonumber\\
&=&\sum_{k}\frac{\hat{P}_{k}^{\beta}}{Z} \log\hat{P}_{k}
\left( \log \hat{P}_{k} -\sum_{m}
\frac{\hat{P}_{m}^{\beta}}{Z}\log \hat{P}_{m} \right)\nonumber\\
&=&\langle \epsilon ^{2}\rangle - \langle \epsilon \rangle^{2},
\end{eqnarray}
where we use Equation (\ref{SelfEne}). 
The Equation (\ref{EnergyFluctuation}) is therefore proved. 

\subsubsection{Pseudo Fisher information and energy fluctuation}
When $\beta$ is a parameter of probability mass function $P$, 
Fisher information $\tilde{I}(\beta)$ is defined in the usual way as
\begin{equation}
\tilde{I}(\beta):= \sum_{k}f_{k}(\beta)
\left( \frac{\partial}{\partial \beta} 
\log f_{k}(\beta) \right)^{2},
\end{equation}
where $f$ is the likelihood function. 
Here, the likelihood function is replaced with the probability function 
and we called the replaced $\tilde{I}(\beta)$ MFE-Fisher information $I(\beta)$.

When it is assumed that the probability functions can be expressed 
by the canonical distributions with parameter $\beta$, 
it is clear that $I(\beta) = \langle \epsilon ^{2}\rangle - \langle \epsilon \rangle^{2}$ 
as follows.
\begin{eqnarray}
I(\beta)
&=& \sum_{k}\frac{\hat{P}_{k}^{\beta}}{Z}
\left( \frac{\partial}{\partial \beta} \log 
\frac{\hat{P}_{k}^{\beta}}{Z} \right)^{2}\nonumber\\
&=& \sum_{k}\frac{\hat{P}_{k}^{\beta}}{Z}
\left( \log \hat{P}_{k} -\frac{1}{Z}
\sum_{m}\hat{P}_{m}^{\beta} \log \hat{P}_{m}
\right)^{2}\nonumber\\
&=& \sum_{k}\frac{\hat{P}_{k}^{\beta}}{Z}
(-\log \hat{P}_{k})^{2} - \left(
\sum_{k}\frac{\hat{P}_{k}^{\beta}}{Z}
(-\log \hat{P}_{k}) \right)^{2}\nonumber\\
&=&\langle \epsilon ^{2}\rangle - \langle \epsilon \rangle^{2}.
\end{eqnarray}
The MFE-Fisher information is therefore identical to 
the energy fluctuation defined in Equation (\ref{EnergyFluctuation}). 

\subsubsection{Other similarities with statistical mechanics}
\label{Sim}
It is noteworthy that MFEE 
has other similarities with thermodynamics and/or statistical mechanics. 
That is, the same relationships exist. 
We list those below: 
\begin{itemize}
\item The following relation is easily derived from the definition 
of partition function $Z$:
\begin{equation}
U=-\frac{\partial}{\partial \beta}\log Z.
\label{U-logZ}
\end{equation}
\item The following relation, known as the Gibbs-Helmholtz relation, 
is derived from Equations (\ref{F-Z}) and (\ref{U-logZ}) as follows:
\begin{equation}
U=\frac{\partial}{\partial \beta}(\beta F).
\label{UbetaF}
\end{equation}
\item The following relation is simply obtained from 
Equations (\ref{FUHBrelation}) and (\ref{UbetaF}):
\begin{equation}
H=\beta^{2} \frac{\partial F}{\partial \beta}.
\end{equation}
\item The energy variance is represented by the second-order differential 
of the partition function for $\beta$ as
\begin{equation}
\langle \epsilon ^{2}\rangle - \langle \epsilon \rangle^{2}
=\frac{\partial^{2}}{\partial \beta^{2}} \log Z.
\end{equation}
\item Thermal capacity is represented by data temperature and energy fluctuation as
\begin{eqnarray}
C&=&\beta^{2}\frac{\partial^{2}}{\partial \beta^{2}}\log Z\nonumber\\
&=&\beta^{2}
(\langle \epsilon ^{2}\rangle - \langle \epsilon \rangle^{2}).
\end{eqnarray}
\end{itemize}

\subsection{Incorporating prior knowledge}
\label{EX}
MFEE can incorporate Bayesian subjective beliefs. 
It can be said that Bayesian subjective probabilities extend counts of events 
to the sum of those counts and prior imaginary counts of the events. 
Our method can therefore easily include subjectivity. 
To achieve this extension, the ML estimators are replaced with Bayesian posterior point 
probabilities such as those 
estimated by the maximizing a posterior (MAP) method, which leads to the Bayesian canonical 
distributions instead of the formula expressed by Equation (\ref{Boltzmann}) as following:
\begin{equation}
P(x)=\frac{\exp(-\beta(-\log P_{Bayes}(x)))}
{\sum_{x'}\exp(-\beta(-\log P_{Bayes}(x')))},\label{BoltzmannBayes}
\end{equation}
where $P_{Bayes}(x)$ is the Bayesian posterior point probability, 
and data temperature is calculated from Equation (\ref{TempFluctuation}) where 
ML estimator is replaced by $P_{Bayes}(x)$. 
In this case, the stronger the subjectivity, the more the data temperature, that is, 
the tempering effect caused by $\beta$ is weaker.

\section{Examples}
\label{Example}
Simulations to demonstrate the robustness for small samples of MFEE, 
in comparison with ML, ME, and Bayesian--Dirichlet estimators 
with Jeffreys' prior, are described. 

$X$ is assumed to have three internal states and four probability mass functions 
with a variety of entropies denoted as $H(X)$ in natural logarithms:
\begin{enumerate}
\item $P(x=0)=0.431, P(x=1)=0.337, P(x=2)=0.232, H(X)=1.07$,
\item $P(x=0)=0.677, P(x=1)=0.206, P(x=2)=0.117, H(X)=0.841$, 
\item $P(x=0)=0.851, P(x=1)=0.117, P(x=2)=0.0320, H(X)=0.498$, 
\item $P(x=0)=0.9898, P(x=1)=0.00810, P(x=2)=0.00210, H(X)=0.0621$. 
\end{enumerate}
Data from each function was sampled, and probabilities were estimated from given data sets with various data sizes. 
In the estimation, ML, ME, and Bayesian--Dirichlet estimation with hyperparameter $\alpha = 1/2$ 
derived from Jeffreys' prior distribution on Dirichlet models~\cite[p. 130]{BayesianChoice} were used. 
The maximizing a posterior (MAP) was used for Bayesian--Dirichlet estimation 
from the viewpoint of point estimations. 
We set, as usual, averaged outputs $\langle X \rangle$ as the constraint in the ME method as following:
$\langle X \rangle := (1/N) \sum_{d=1}^{N} X_{d}$, where $X_{d}$ denotes $d$-th sample's output 
and $N$ denotes a sample size. 
After that, true and estimated probabilities were compared by using 
Kullback-Leibler (KL) divergence~\cite{KL} as a metric, 
which has the following form: 
\begin{equation}
D(P(X)\,||\,P_{\rm{e}}(X))=\sum_{x}P(x)\log\frac{P(x)}{P_{\rm{e}}(x)},
\end{equation}
where $P(X)$ is the true distribution, and $P_{\rm{e}}(X)$ is the distribution estimated 
by ML, ME, MAP, or MFEE. 
For avoiding zero probabilities, probabilities were smoothed by adding $0.0001$ to the counts.

\begin{figure*}[t]
\leavevmode
\begin{center}
\vspace{-6mm}
  \begin{tabular}{c c}
    \includegraphics[width=2.65in]
                     {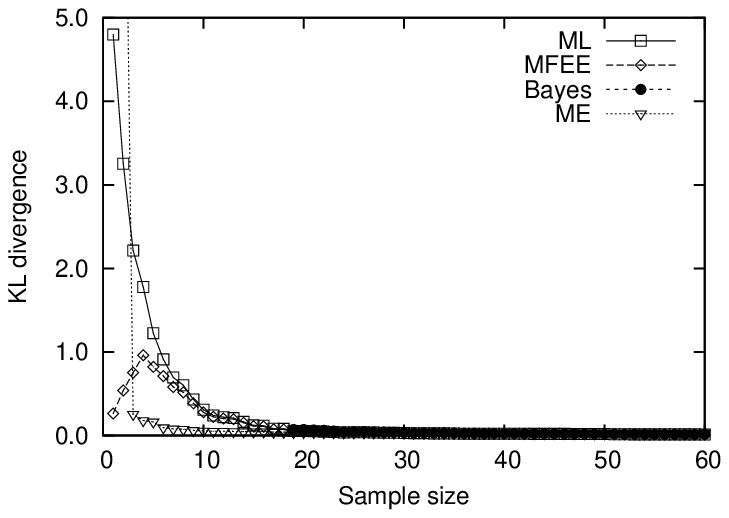} &
    \includegraphics[width=2.65in]
                     {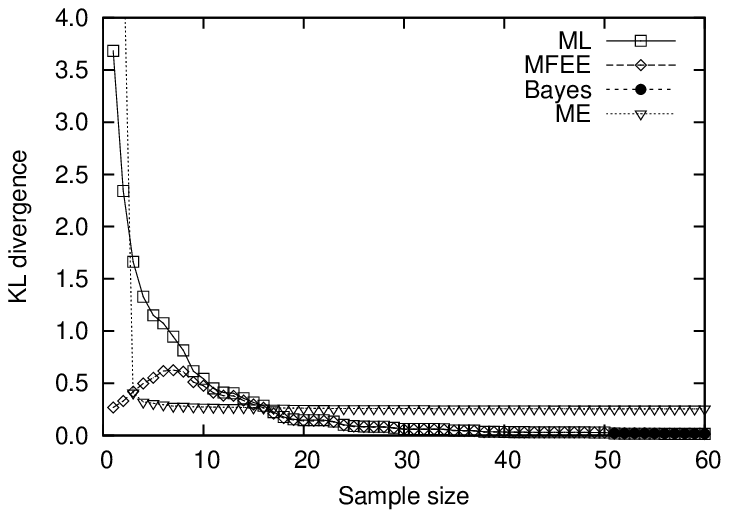} \\
                     {\tiny }\\
                     (a): $H=1.07$&
                     (b): $H=0.841$\\
                     \\
                     \\
    \includegraphics[width=2.65in]
                     {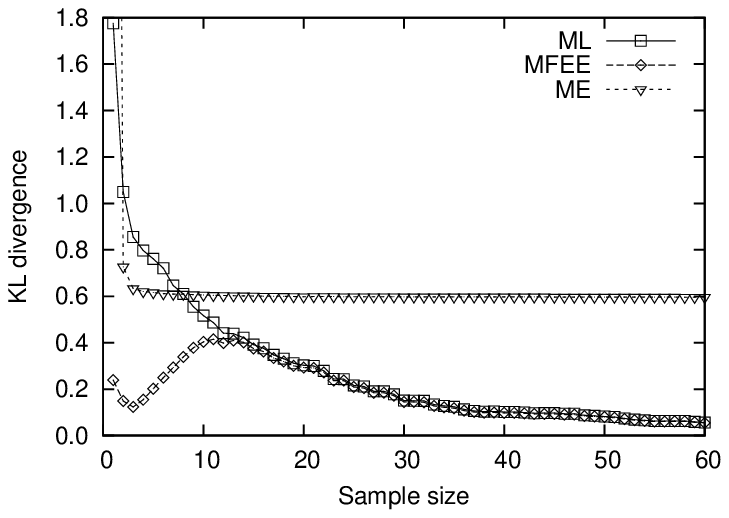} &
    \includegraphics[width=2.65in]
                     {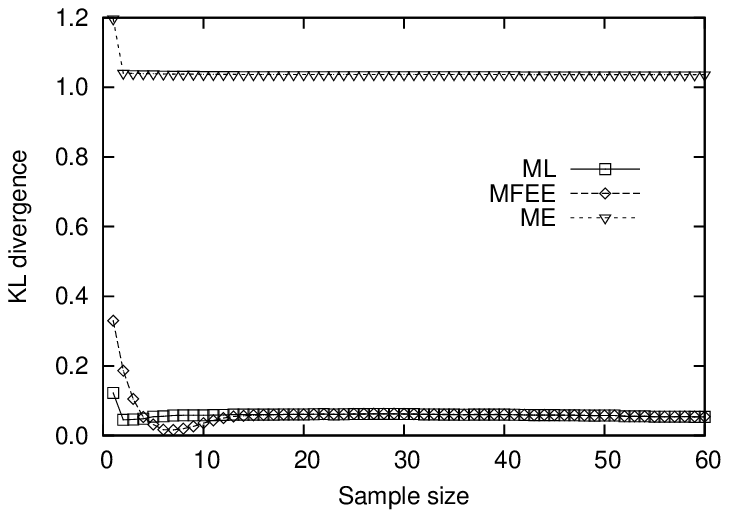} \\
                     {\tiny  }\\
                     (c): $H=0.498$&
                     (d): $H=0.0621$\\
                     {\tiny  }\\
		     {\tiny  }\\
  \end{tabular}
\caption{KL divergences between true probability mass functions and probability mass 
functions estimated by using maximum likelihood (ML), MFEE, 
maximum a posterior with Jeffreys' prior (Bayes) and maximum entropy (ME). 
The horizontal axes denote sample sizes. 
$H$ denotes Shannon entropy in natural logarithms.}
  \label{EstimationExamples}
\end{center}
\end{figure*}

The KL divergences are shown in Fig.~\ref{EstimationExamples}, where they are averaged values 
from 100 samples at each sample size from identical distributions. 
It can be seen that the ML estimators are inferior to MFEE due to overfitting, 
except for the distribution having very small entropy. 
Even the degree of superiority of the ML estimation in (d) is 
relatively smaller than that of inferiority in other distributions. 
The ME estimators showed the opposite behaviors to the ML, and showed some relatively poor 
results in large-sample regions. 
ML methods tend to fit data and then can more accurately estimate distributions 
with very low entropies than others in small sample cases. 
For example, if a true entropy equals zero, 
ML method can estimate the exact true distribution from only one sample. 
On the other hand, ME methods tend to increase entropies 
and then can with high entropies such as the uniform distributions. 
Hence, the ML has tendency of overfitting and the ME has that of underfitting 
in the view of misestimation. 
Even so MFEE showed relative stability in both sample sizes and distributions. 
It indicates effectiveness of MFEE method, as was defined so as to 
incorporate characteristics of both the ME and ML.
Not all values could be estimated by the MAP at each sample size for the following 
sample sizes ($N$): $N\leq 19$ in (a), $N\leq 50$ in (b), $N\leq 103$ in (c), and $N>500$ in (d), 
where the size decreased in response to the values of entropies for each distribution. 
This is because even the posterior probability distributions were improper distributions, 
which have been pointed out as a problem of Bayesian statistics~\cite{KassWassermanSelection}. 
In these ranges, the averaged points about the MAP were not plotted in Fig.~\ref{EstimationExamples}. 
Our method always provides the values for any sample size and showed effectiveness for avoiding 
overfitting (at least to some extent). 

\section{Discussion}
\label{Discussion}

\subsection{Relation to classical and Jaynes' approach}
The classical approach to probability estimation based on frequentism is included in our method. 
This approach can be considered as a method where $\beta$ is assumed to be $1$ or nearly $1$ 
in our MFE-based method. That is, the approach can be said to be a \textit{zero-temperature} 
or a \textit{low-temperature} approximation of our method. 

MFEE suggests a new interpretation of probability, in which sample size is not 
explicitly included, unlike both the classical and Bayesian approaches. 
Although an equivalent sample size can be calculated from a probability estimated by MFEE, 
the calculated value is no more than the one as interpreted in the language of frequentism. 

Jaynes' maximum entropy (ME) methods are well known as the least biased inference methods. 
However, the constraints on which ME methods are based may not be reliable for small samples and then 
may be biased. On the other hand, MFEE corrects even such biases using temperature. 
Moreover, ME methods seemed to fail in estimation from large size-samples in our simulations, which 
implies that ME methods do not fully take advantage of information from available data. 

\subsection{Relation to Bayesian approach}
Our approach is quite different from Bayesian approaches when prior knowledge is not available 
or should be excluded. MFEE assumes that a physics-like mechanism determines optimal estimation, 
while the Bayesian approach assumes noninformative prior distributions unrelated to the optimal estimation. 
In addition, the former puts optimal entropy at the center of the method, which seems desirable 
because statistical inference aims to get optimal useful information from data. 

In hierarchical Bayesian models, the hyperparameters of prior distributions 
are often determined by maximizing marginal distributions, 
which are called the empirical Bayesian methods. 
Our method is not suitable for these models because hyperparameters are not ones of noninformative priors and 
can be interpreted as \textit{additive parameters} that complement incompleteness of structures of the models. 
A similar situation occurs in Bayesian network classifiers (BNCs)~\cite{FGG}, 
as mentioned in our previous work~\cite{isozakiICTAI,Isozaki_IJAIT}. 
BNCs are being developed in the machine learning domain for classification tasks and 
are generalized from well-known naive Bayes classifiers. 
In the case of BNCs, it is known that their conditional probabilities play a part in complementing 
inaccuracies of estimated network structures~\cite{Jing}. 

\section{Summary}
\label{Conclusion}
Based neither on frequentism nor Bayesianism, a robust method of the probability estimation-based 
on both thermodynamics and information theory-for discrete-random-variable systems was developed. 
The core of the method is the intent to obtain optimized entropy explicitly, namely, obtaining optimized 
information from available limited data. 
The theory introduces two new quantities: \textit{information energy} and \textit{data temperature}. 
\textit{Free energy} is defined by using these quantities. 
The minimum free energy principle for inference, which unifies the maximum likelihood and maximum entropy 
principles with the above quantities, is adopted. 
The theory has advantages over frequentism because of it is more robust for small sample size and over 
Bayesianism because it does not use prior/posterior distributions when no prior knowledge is available, 
where prior biases are regarded as not completely excluded. 
The effectiveness of the method in terms of robustness 
was demonstrated by simulation studies on point estimation for single variable systems 
with various entropies. 

\section*{Acknowledgments}
The author would like to thank Mario Tokoro and Hiroaki Kitano 
of Sony Computer Science Laboratories, Inc. for their support. 

\bibliographystyle{apalike}
\bibliography{BetaLearning}

\begin{thebibliography}{}

\bibitem[Basu et~al., 1998]{BasuBetaLikelihood}
Basu, A., Harris, I.~R., Hjort, N.~L., and Jones, M.~C. (1998).
\newblock Robust and efficient estimation by minimising a density power
  divergence.
\newblock {\em Biometrika}, 85:549--559.

\bibitem[Berger et~al., 1996]{MEtoNLP}
Berger, A.~L., Pietra, S.~D., and Pietra, V.~D. (1996).
\newblock A maximum entropy approach to natural language processing.
\newblock {\em Computational Linguistics}, 22(1):39--71.

\bibitem[Callen, 1985]{CallenThermodynamics}
Callen, H.~B. (1985).
\newblock {\em Thermodynamics and An Introduction to Thermostatistics}.
\newblock John Wiley \& Sons, Hoboken, NJ, second edition.

\bibitem[Caticha and Preuss, 2004]{MEpriorPRE04}
Caticha, A. and Preuss, R. (2004).
\newblock Maximum entropy and {B}ayesian data analysis: Entropic prior
  distributions.
\newblock {\em Physical Review E}, 70:046127.

\bibitem[Cover and Thomas, 2006]{CoverThomas}
Cover, T.~M. and Thomas, J.~A. (2006).
\newblock {\em Elements of Information Theory}.
\newblock John Wiley \& Sons, Hoboken, NJ, second edition.

\bibitem[Dose, 2003]{BayesInPhysDose}
Dose, V. (2003).
\newblock Bayesian inference in physics: Case studies.
\newblock {\em Reports on Progress in Physics}, 66:1421.

\bibitem[Friedman et~al., 1997]{FGG}
Friedman, N., Geiger, D., and Goldszmidt, M. (1997).
\newblock {B}ayesian network classifiers.
\newblock {\em Machine Learning}, 29(2-3):131--163.

\bibitem[H\'{a}jek, 1997]{HajekArguments1997}
H\'{a}jek, A. (1997).
\newblock ''mises redux'' --redux: Fifteen arguments against finite
  frequentism.
\newblock {\em Erkenntnis}, 45:209--227.

\bibitem[Hofmann, 1999]{Hofmann99UAI}
Hofmann, T. (1999).
\newblock Probabilistic latent semantic analysis.
\newblock In {\em Proc. of Conference on Uncertainty in Artificial Intelligence
  (UAI-99)}, pages 289--296.

\bibitem[Huscroft et~al., 2000]{ME_Huscroft2000}
Huscroft, C., Gass, R., and Jarrell, M. (2000).
\newblock Maximum entropy method of obtaining thermodynamical properties from
  quantum {M}onte {C}arlo simulations.
\newblock {\em Physical Review B}, 61:9300.

\bibitem[Irony and Singpurwalla, 1997]{NonInformativeDialogue}
Irony, T.~Z. and Singpurwalla, N.~D. (1997).
\newblock Non-informative priors do not exist: A dialogue with {J}os\'{e}
  {M}.~{B}ernardo.
\newblock {\em Journal of Statistical Planning and Inference}, 65:159--189.

\bibitem[Isozaki et~al., 2008]{isozakiICTAI}
Isozaki, T., Kato, N., and Ueno, M. (2008).
\newblock Minimum free energies with ``data temperature'' for parameter
  learning of {B}ayesian networks.
\newblock In {\em Proc. of {IEEE} International Conference on Tools with
  Artificial Intelligence (ICTAI-08)}, pages 371--378.

\bibitem[Isozaki et~al., 2009]{Isozaki_IJAIT}
Isozaki, T., Kato, N., and Ueno, M. (2009).
\newblock ``{D}ata temperature'' in minimum free energies for parameter
  learning of {B}ayesian networks.
\newblock {\em International Journal on Artificial Intelligence Tools},
  18(5):653--671.

\bibitem[Jaynes, 1957]{Jaynes57I}
Jaynes, E.~T. (1957).
\newblock Information theory and statistical mechanics.
\newblock {\em Physical Review}, 106(4):620--630.

\bibitem[Jeffreys, 1961]{JeffreysBook}
Jeffreys, H. (1961).
\newblock {\em Theory of Probability}.
\newblock Oxford University Press, NY, third edition.

\bibitem[Jing et~al., 2005]{Jing}
Jing, Y., Pavlovi\'{c}, V., and Rehg, J.~M. (2005).
\newblock Efficient discriminative learning {B}ayesian network classifier via
  boosted augmented naive {B}ayes.
\newblock In {\em Proc. of International Conference on Machine Learning
  (ICML-05)}, pages 369--376.

\bibitem[Jones et~al., 2001]{JonesBiometrika2001}
Jones, M.~C., Hjort, N.~L., Harris, I.~R., and Basu, A. (2001).
\newblock A comparison of related density-based minimum divergence estimators.
\newblock {\em Biometrika}, 88(3):865--873.

\bibitem[Kass and Wasserman, 1996]{KassWassermanSelection}
Kass, R.~E. and Wasserman, L. (1996).
\newblock The selection of prior distributions by formal rules.
\newblock {\em Journal of the American Statistical Association},
  91(435):1343--1370.

\bibitem[Kittel and Kroemer, 1980]{Kittel}
Kittel, C. and Kroemer, H. (1980).
\newblock {\em Thermal Physics}.
\newblock W.~H.~Freeman, San Francisco, CA.

\bibitem[Kullback and Leibler, 1951]{KL}
Kullback, S. and Leibler, R.~A. (1951).
\newblock On information and sufficiency.
\newblock {\em Annals of Mathematical Statistics}, 22(1):79--86.

\bibitem[LeCun and Huang, 2005]{YannLeCun_AISTATS05}
LeCun, Y. and Huang, F.~J. (2005).
\newblock Loss functions for discriminative training of energy-based models.
\newblock In {\em Proc. of International Workshop on Artificial Intelligence
  and Statistics (AISTATS-05)}, pages 206--213.

\bibitem[Pereira et~al., 1993]{Pereira93ACL}
Pereira, F., Tishby, N., and Lee, L. (1993).
\newblock Distributional clustering of {E}nglish words.
\newblock In {\em Proc. of Annual Meeting on Association for Computational
  Linguistics (ACL-93)}, pages 183--190.

\bibitem[Robert, 2007]{BayesianChoice}
Robert, C.~P. (2007).
\newblock {\em The Bayesian Choice: From Decision-Theoretic Foundations to
  Computational Implementation}.
\newblock Springer-Verlag, New York, NY, second edition.

\bibitem[Shannon, 1948]{Shannon}
Shannon, C.~E. (1948).
\newblock A mathematical theory of communication.
\newblock {\em Bell Systems Technical Journal}, 27:379--423,623--656.

\bibitem[Ueda and Nakano, 1995]{UedaDAEM}
Ueda, N. and Nakano, R. (1995).
\newblock Deterministic annealing variant of the {EM} algorithm.
\newblock In {\em Proc. of Advances in Neural Information Processing Systems 7
  (NIPS 7)}, pages 545--552.

\bibitem[Watanabe et~al., 2009]{KWataVFEforBN}
Watanabe, K., Shiga, M., and Watanabe, S. (2009).
\newblock Upper bound for variational free energy of {B}ayesian networks.
\newblock {\em Machine Learning}, 75(2):199--215.

\bibitem[Windham, 1995]{Windham1995Robust}
Windham, M.~P. (1995).
\newblock Robustifying model fitting.
\newblock {\em Journal of the Royal Statistical Society {B}}, 57(3):599--609.

\end{thebibliography}
\end{document}